\documentclass[lettersize,journal]{IEEEtran}
\usepackage{amsmath,amsfonts}
\usepackage[ruled,linesnumbered,noend]{algorithm2e}
\usepackage{array}
\usepackage[caption=false,font=normalsize,labelfont=sf,textfont=sf]{subfig}
\usepackage{textcomp}
\usepackage{stfloats}
\usepackage{url}
\usepackage{verbatim}
\usepackage{graphicx}
\usepackage{cite}
\usepackage[colorlinks,urlcolor=black,linkcolor=blue,citecolor=blue]{hyperref}
\usepackage{booktabs}
\usepackage{multirow}
\usepackage{makecell}
\usepackage{xcolor}
\usepackage{enumitem}
\usepackage{wrapfig}
\usepackage[font=small]{caption}
\usepackage[capitalize,noabbrev]{cleveref}

\DeclareMathOperator*{\argmin}{arg\,min}
\DeclareMathOperator*{\argmax}{arg\,max}
\newcommand{\AP}[1]{AP\textsubscript{#1}}
\newcommand{\redText}[1]{\textcolor{red}{#1}}

\begin{document}

\title{DIVE: Inverting Conditional Diffusion Models\\for Discriminative Tasks}

\author{Yinqi~Li,
        Hong~Chang,~\IEEEmembership{Member,~IEEE,}
        Ruibing~Hou,~\IEEEmembership{Member,~IEEE,}\\
        Shiguang~Shan,~\IEEEmembership{Fellow,~IEEE,}
        and~Xilin~Chen,~\IEEEmembership{Fellow,~IEEE}
\thanks{Y. Li, H. Chang (corresponding author), S. Shan and X. Chen are with Key Laboratory of Intelligent Information Processing, Institute of Computing Technology, Chinese Academy of Sciences, Beijing, 100190, China, and University of Chinese Academy of Sciences, Beijing 100049, China.}
\thanks{R. Hou is with Key Laboratory of Intelligent Information Processing, Institute of Computing Technology, Chinese Academy of Sciences, Beijing, 100190, China.}
\thanks{E-mail: yinqi.li@vipl.ict.ac.cn, \{changhong, houruibing, sgshan, xlchen\}@ict.ac.cn}
\thanks{Preprint. © 2025 IEEE. Personal use of this material is permitted.  Permission from IEEE must be obtained for all other uses, in any current or future media, including reprinting/republishing this material for advertising or promotional purposes, creating new collective works, for resale or redistribution to servers or lists, or reuse of any copyrighted component of this work in other works.}}

\markboth{Preprint}%
{Li \MakeLowercase{\textit{et al.}}: DIVE: Inverting Conditional Diffusion Models for Discriminative Tasks}


\maketitle

\begin{abstract}
Diffusion models have shown remarkable progress in various generative tasks such as image and video generation.
This paper studies the problem of leveraging pretrained diffusion models for performing discriminative tasks.
Specifically, we extend the discriminative capability of pretrained frozen generative diffusion models
from the classification task~\cite{li2023your, clark2023text} to the more complex object detection task,
by ``inverting'' a pretrained layout-to-image diffusion model.
To this end, a gradient-based discrete optimization approach for replacing the heavy prediction enumeration process, 
and a prior distribution model for making more accurate use of the Bayes' rule,
are proposed respectively.
Empirical results show that this method is on par with basic discriminative object detection baselines on COCO dataset. 
In addition, our method can greatly speed up the previous diffusion-based method~\cite{li2023your, clark2023text} for classification without sacrificing accuracy.
Code and models are available at \url{https://github.com/LiYinqi/DIVE}.
\end{abstract}


\section{Introduction}
\label{sec:intro}

\IEEEPARstart{R}{ecently}, generative models such as diffusion models~\cite{sohl2015deep, ho2020denoising, song2021scorebased, dhariwal2021diffusion, rombach2022high, saharia2022photorealistic}, 
autoregressive models~\cite{van2017neural, razavi2019generating, esser2021taming, ramesh2021zero},
and generative adversarial networks (GANs)~\cite{goodfellow2014generative, brock2019large, karras2019style} have gained increasing interest in the research community due to their ability to synthesize photorealistic images.
This remarkable generation ability implies that these methods could accurately model the data distribution and learn effective image representations.

Leveraging pretrained generative models, 
some works~\cite{he2023is, sariyildiz2023fake, azizi2023synthetic, wu2023diffumask, nguyen2023dataset, chen2024geodiffusion, fang2024data} synthesize a set of training samples and then use the synthetic training set to train corresponding models for discriminative tasks.
Some other works~\cite{baranchuk2022labelefficient, xu2023ODISE, yang2023diffusion, li2023dreamteacher, zhao2023unleashing, kondapaneni2023textimage, xu2024diffusion} utilize the trained generative model to extract features and learn discriminative heads upon.
Notably, both of these two families of approaches still rely on and train discriminative models to perform discriminative tasks.
Instead, in this work, we study whether purely pretrained generative models could accomplish discriminative tasks.
That is, we focus on directly converting pretrained generative models to ``discriminative models'', without finetuning the model parameters or training additional discriminative heads.
We believe that such a study could more intrinsically reveal the discriminative ability of pretrained generative models.

\begin{figure}[t]
  \centering
  \begin{tabular}{@{}c@{}}
      \includegraphics[width=.9\linewidth]{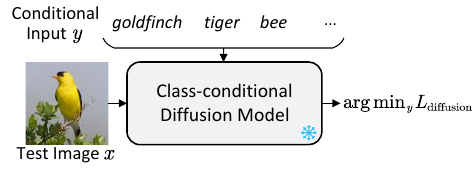}\\[-0.02in]
      \small (a) Enumeration-based Diffuser Classifier~\cite{li2023your, clark2023text}.\\
  \end{tabular}
  \vspace{0.09in}
  
  \begin{tabular}{@{}c@{}}
      \includegraphics[width=.9\linewidth]{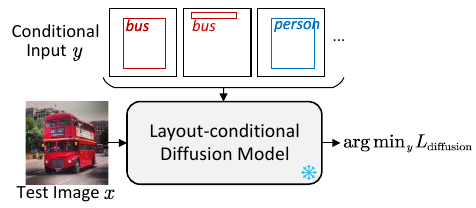}\\[-0.02in]
      \small (b) Directly extending (a) to the object detection task.\\
  \end{tabular}
  \vspace{0.09in}
  
  \begin{tabular}{@{}c@{}}
      \hspace{-1.5mm}\includegraphics[width=.903\linewidth]{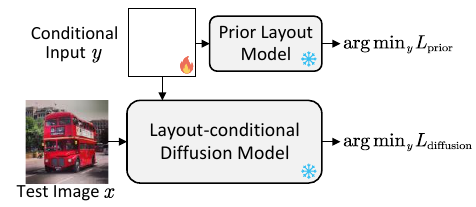}\\[-0.02in]
      \small (c) Our proposed optimization-based prior-integrated framework.\\
  \end{tabular}
  \vskip -0.04in
  \caption{Repurposing pretrained conditional diffusion models for discriminative tasks without tuning model parameters.
  $L_{\text{diffusion}}$ and $L_{\text{prior}}$ represent the training loss of the diffusion model~\cite{ho2020denoising}, which is the lower bound of the modeled data distribution $p(x|y)$ and $p(y)$.
  }
  \label{fig:teaser}
  \vskip -0.23in
\end{figure}

\vspace{0.02in}
The classical way of such a generative-to-discriminative conversion is through the Bayes' rule, which accomplishes the discriminative task $\argmax_y p(y|x)$ by calculating $\argmax_y p(x|y) p(y)$, where $y$ is the task label and $x$ is the input image~\cite{ng2001discriminative}.
Based on this approach, recently, \cite{li2023your, clark2023text} convert pretrained class- (and text-) conditional diffusion models, a family of likelihood-based generative models, to generative classifiers.
For a $C$-way classification task under the assumption of a uniform prior $p(y)$, their solution is to query the model $C$ times to obtain the likelihood of each class and then pick the highest one, as shown in \cref{fig:teaser}(a).

In this paper, we explore how such an inversion approach can be applied to more complex discriminative tasks, such as object detection, as shown in \cref{fig:teaser}(b).
We argue that this idea is interesting because it
provides a purely generative-based approach to the complex object detection task,
helping researchers understand pretrained diffusion models from the perspective of discriminative ability, 
and by-productively offering a novel evaluation metric for diffusion models.
Meanwhile, previous diffusion-based classification methods~\cite{li2023your, clark2023text} cannot be directly extended due to the following challenges:
(1) the prior label distribution of detection datasets, i.e., object layouts, is not uniform and thus can not be ignored;
(2) more critically, it is impossible to enumerate all potential detection labels for a test image, as the combinations of objects and their locations are vast and almost countless.

To solve the first problem,
we train a diffusion model to model the prior $p(y)$. Together with the pretrained conditional diffusion model $p(x|y)$, we obtain the discriminative prediction $p(y|x)$ through posterior maximization.
For the second problem, rather than testing each potential label $y$ as the previous work does, we propose to treat $y$ as a to-be-optimized result and find it by maximizing the posterior objective. 
Specifically, we freeze the parameters of the trained diffusion models while allowing the conditional input to be a learnable embedding, and use the gradient method to optimize it.
The overall framework is illustrated in \cref{fig:teaser}(c) and described in detail in \cref{sec:method}.
We name this method Diffusion model InVErsion (DIVE) since it inverts a $y$-to-$x$ model to $x$-to-$y$. 

We mainly introduce and evaluate our method in the context of the standard object detection task.
Experiments on COCO~\cite{lin2014microsoft} show that using our method to invert a diffusion model achieves competitive results against basic discriminative detectors such as Faster R-CNN~\cite{ren2015faster}.
In addition, we apply our optimization-based method to the image classification task.
Compared to the previous enumeration-based method~\cite{li2023your}, our method significantly improves the speed while maintaining accuracy.
Although there are still performance and efficiency gaps between our approach and the discriminative models,
we hope that this work could help researchers further recognize the discriminative ability of pretrained generative models, and could serve as a potential novel solution for complex discriminative tasks.

Our contributions can be summarized as follows:
\begin{itemize}[leftmargin=*]
\item We extend the paradigm of inverting diffusion models for discriminative tasks from classification to object detection.
To the best of our knowledge, this is the first work to invert an image-generation model for the object detection task.

\item We propose learning modules to make the inversion paradigm theoretically more accurate and implementationally more feasible for the more complex detection task.
    
\item We empirically show that our method not only achieves competitive results with basic discriminative detectors but also performs faster than enumeration-based generative classifiers.

\end{itemize}

\section{Related Work}
\label{sec:rw}

\subsection{Diffusion Models}
Diffusion models are a type of generative models that approximate data distribution by gradually adding noise to the data in a forward process and learning to reverse this process through a Markov chain to generate new data~\cite{sohl2015deep,ho2020denoising}.
Diffusion models have gained superior attention in various generative tasks, including image generation~\cite{dhariwal2021diffusion, rombach2022high, saharia2022photorealistic, 10589534, 10528891}, image editing~\cite{meng2022sdedit, 10261222, 10480591}, and video generation~\cite{ho2022video, 10321681, 10420468}, etc., 
due to their ability of synthesizing high-fidelity contents.

\subsection{Generative Models for Discriminative Tasks}
Driven by the superior content generation capacity, recently, there has been an increase in using generative models to assist or perform discriminative tasks.
Typical works can be categorized as:

(1) Generating images to train discriminative models~\cite{he2023is, ni2023imaginarynet, sariyildiz2023fake, azizi2023synthetic, wu2023diffumask, you2023diffusion, dunlap2023diversify, nguyen2023dataset, yang2023freemask, karazija2023diffusion, xie2023mosaicfusion, suri2024gendet, chen2024geodiffusion, fang2024data}; 

(2) Extracting representations from pretrained generative models followed by learnable discriminative heads~\cite{baranchuk2022labelefficient, xu2023ODISE, yang2023diffusion, li2023dreamteacher, zhao2023unleashing, kondapaneni2023textimage, xu2024diffusion, li2023open, mukhopadhyay2023textfree, dong2024bridging, patni2024ecodepth}; 

(3) Generating task labels conditioned on images with architecture modification and additional training~\cite{amit2021segdiff, chen2023generalist, chen2023diffusiondet, duan2023diffusiondepth, ji2023ddp, lee2023exploiting, ke2023repurposing, 10520935}; 

(4) Leveraging the attention maps~\cite{vaswani2017attention} from the model to accomplish segmentation-related tasks~\cite{wu2023diffumask, tian2023diffuse, pnvr2023ld, wang2023diffusion, xiao2023text}, which are mainly based on text-to-image diffusion models~\cite{rombach2022high, saharia2022photorealistic}.

Among these works, (4) is only applicable under specific architectures, (1)-(3) require additional training, and some of them even contain discriminative models.
In sum, they do not intrinsically reveal the discriminative ability of pretrained image-generation models.

Unlike all these works, some recent methods directly convert image-generation diffusion models to perform discriminative tasks using Bayes' rule~\cite{ng2001discriminative}.
\cite{li2023your, clark2023text} build standard (or zero-shot) image classifiers by converting class-~\cite{peebles2023scalable} (or text-~\cite{rombach2022high, saharia2022photorealistic}) conditional diffusion models, 
and \cite{zimmermann2021score, chen2023robust, chen2024diffusion} focus on building robust diffusion classifiers.
On the other hand, \cite{krojer2023diffusion, rambhatla2023selfeval} extend this paradigm to image-text-matching tasks and use it to evaluate text-to-image generation models.
In this work, we further explore whether other more complex discriminative tasks like object detection, i.e., without a set of predefined possible class~\cite{li2023your, clark2023text} or text~\cite{krojer2023diffusion, rambhatla2023selfeval} labels for the test image, can be accomplished under the model inversion paradigm.

\subsection{Generative Model Inversion}
Generative model inversion usually refers to finding the initial noise representation of a given image, with which one can reconstruct that image using the generative model~\cite{zhu2016generative, xia2022gan}.
This is often applied for editing real images, by manipulating the inverted noise representations~\cite{zhu2016generative, xia2022gan, tumanyan2023plug, cao2023masactrl}.
For GANs, the inversion can be accomplished through 
optimization-based approaches that optimize the initial noise by minimizing the reconstruction loss~\cite{creswell2018inverting, abdal2019image2stylegan, zhu2020improved}; 
encoder-based approaches that learn an encoder mapping images to latent noises~\cite{Richardson_2021_CVPR, tov2021designing, alaluf2022hyperstyle, 9380493}; 
or combination of the two approaches~\cite{zhu2016generative, zhu2020domain}.
For diffusion models, the inversion can be done through a generalized deterministic forward diffusion process proposed in DDIM~\cite{song2021denoising}.
Following improvement works include exacter and faster inversions~\cite{wallace2023edict, zhang2023exact} and extension to text-conditional models~\cite{mokady2023null, miyake2023negative}. 

Whereas these works invert an image to the noise space for the purpose of exact reconstruction,
this work inverts diffusion models to the conditional input space to produce discriminative predictions.
Recently, \cite{gal2023an, wei2023diffusion, mahajan2023prompting} invert diffusion models to the conditional input space, but their goals are to invert the concept~\cite{gal2023an} or the full content~\cite{wei2023diffusion, mahajan2023prompting} of the given image to texts, and use the inverted texts to generate new images, i.e., still towards the reconstruction goal which is different from ours of performing discriminative task.

\section{Method}
\label{sec:method}

In this section, we first review the background of conditional diffusion models and then introduce our method of repurposing these models to perform discriminative tasks in the context of object detection.

\begin{figure*}[t]
  \centering
  \vskip -0.05in
  \hspace{.5cm}\includegraphics[width=0.81\textwidth]{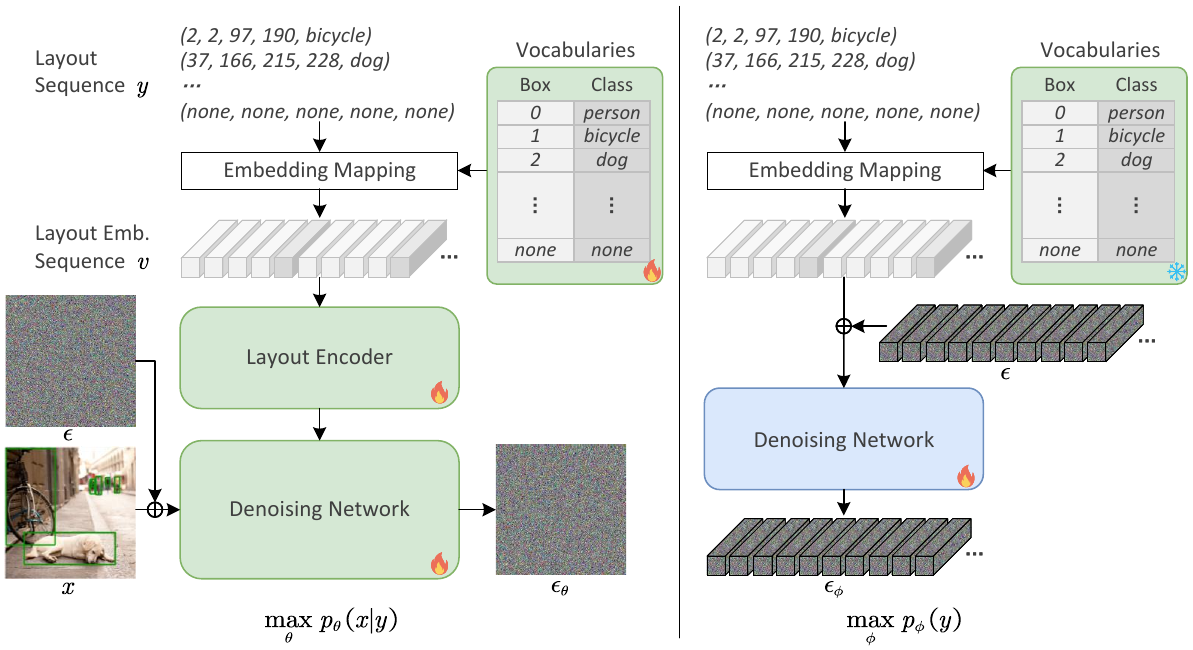}
  \vskip -0.06in
  \caption{Illustration of the \textbf{training} framework of layout-conditional image generation model (left) and prior layout model (right). Bounding boxes in image $x$ are for visualization only.}
  \vskip -0.2in
  \label{fig:dm_training}
\end{figure*}

\subsection{Conditional Diffusion Model Preliminaries}

\subsubsection{Diffusion Model}
Diffusion models~\cite{sohl2015deep, ho2020denoising, song2021scorebased} are likelihood-based generative models trained to reverse a diffusion process.
The forward process, defined as a Markov chain structure, gradually adds noise to the clean image $x_0$ for a number of timesteps $t = 1 \cdots T$~\cite{sohl2015deep}.
During the reverse process, the model learns to predict the amount of noise $\epsilon$ added to the image~\cite{ho2020denoising}, namely $\epsilon_\theta(x_t, t)$, or $\epsilon_\theta(x_t, t, v_\theta(y))$ for conditional diffusion models, where $x_t$ is the noisy image and $y$ is the conditioning input such as class label, and $v_\theta(\cdot)$ maps $y$ to embeddings.
Specifically, the model can be trained by optimizing a variational bound of the conditional log-likelihood~\cite{ho2020denoising}:
\begin{equation}\label{eq:cond_dm_train}
    \max_\theta \log p_\theta(x|y) 
    \approx
    \min_\theta \mathbb{E}_{\epsilon \sim N(0,1), t}{
        \left[ || \epsilon - \epsilon_\theta(x_t, t, v_\theta(y)) ||_2^2 \right],
    }
\end{equation}
with $x, y$ from training data (we use $x$ to represent $x_0$ for simplicity) and $t$ sampled uniformly from $\{ 1 \cdots T \}$.

\vspace{0.02in}
\subsubsection{Conditional Branch}
The conditional branch encodes the input $y$ into the denoising network, facilitating controllable image synthesis, such as class-to-image, text-to-image, and layout-to-image generation.
The encoding of class labels can be easily implemented using a learnable embedding layer that maps integer class indices to embedding vectors, as a branch of works do~\cite{mirza2014conditional, brock2019large, dhariwal2021diffusion, rombach2022high, peebles2023scalable}.

The Latent Diffusion Model (LDM)~\cite{rombach2022high} can deal with a variety of conditioning inputs, including image layouts such as detection labels. 
Building on this capability, we propose a model inversion approach in this paper.
As illustrated in \cref{fig:dm_training}~(left),
similar to the encoding process of class labels, each bounding box $(m_{min}, n_{min}, m_{max}, n_{max})$ with the corresponding class $c$ in an image is firstly mapped to embedding vectors through learnable vocabularies.
These box and class embeddings form a sequence.
Since different images usually contain varying numbers of objects, the sequences are padded with [none] embeddings to achieve a fixed length.
The sequence is then processed by a transformer encoder~\cite{vaswani2017attention} and later integrated into the main denoising network through cross-attention layers~\cite{vaswani2017attention}.
The vocabularies and the transformer encoder are trained as components of the overall diffusion model.

\subsection{Conditional Diffusion Model Inversion for Discriminative Task}
\label{sec:method_inversion}

Given the pretrained conditional diffusion models $p_\theta(x|y)$, following \cite{li2023your, clark2023text}, we use them to perform discriminative tasks using the Bayes' rule:
\begin{equation}\label{eq:gen_to_dis}
    \argmax_y p_\theta(y|x) = \argmax_y p_\theta(x|y) p(y).
\end{equation}

A theoretical issue arises: the conditional diffusion models only capture the conditional likelihood $p_\theta(x|y)$ without considering the prior $p(y)$.
In the case of a class-conditional diffusion model trained on balanced classification datasets like ImageNet-1k~\cite{russakovsky2015imagenet}, 
it is reasonable to disregard $p(y)$ in Eq.~\eqref{eq:gen_to_dis} when converting the model to a ``Diffusion Classifier'' by calculating each class's posterior~\cite{li2023your}, 
because the prior is a uniform distribution (i.e., the same for each possible class).
Unfortunately, for the object detection task, such a uniform prior distribution does not hold at all.
For example, rare layouts, such as a bicycle on top of a person or things with unusual aspect ratios, should have a relatively lower prior probability.
Thus, $p(y)$ could not be ignored when using Eq.~\eqref{eq:gen_to_dis} to obtain diffusion detectors.

\begin{figure*}
  \centering
  \vskip -0.07in
  \hspace{.5cm}\includegraphics[width=0.81\textwidth]{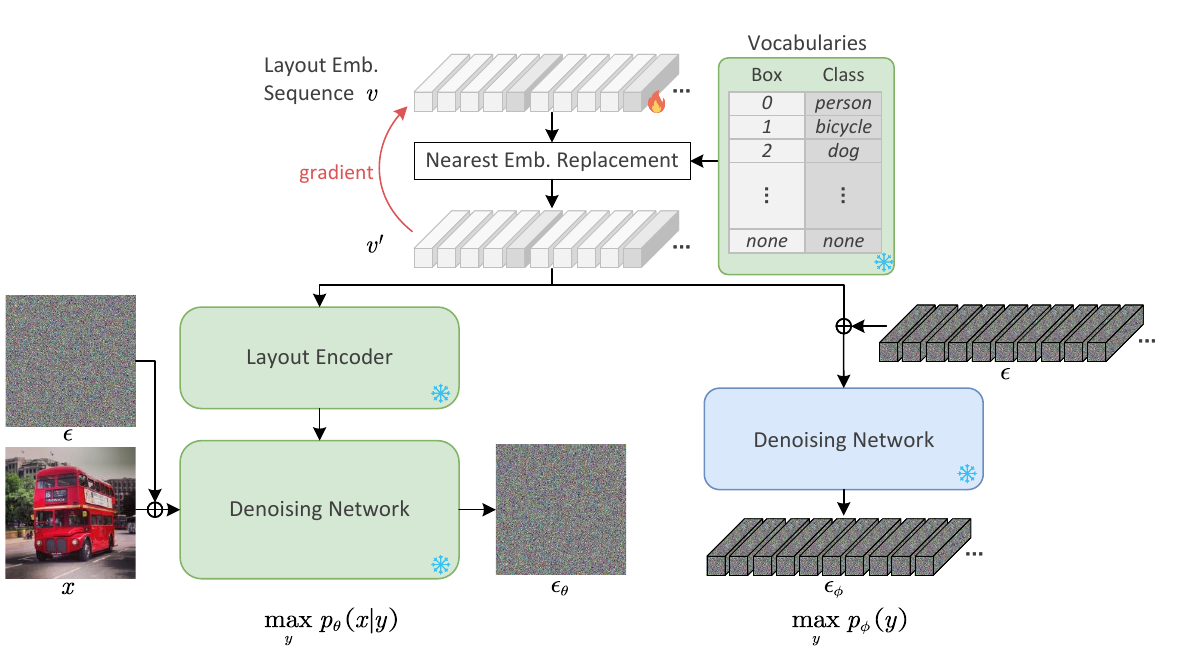}
  \caption{\textbf{Using trained} layout-conditional image generation model and the prior layout model for object detection.}
  \label{fig:dm_inv}
  \vskip -0.15in
\end{figure*}

\vskip 0.02in
\subsubsection{Prior Modeling}
To solve the above issue, we propose to model the underlying prior distribution of the labels in the training dataset.
Ideally, the learned model could tell a reasonable $p(y)$ value for any detection layout $y$.
We construct a diffusion model $p_\phi(y)$ to model the prior label distribution.
Unlike conditional image generation models, this prior model is a label generation model operating within the label space.
Specifically for object detection, the input is a sequence of noised layout embeddings, and the model is trained to predict the added noise sequence:
\begin{equation}\label{eq:label_dm_train}
    \max_\phi \log p_\phi(y) 
    \approx
    \min_\phi \mathbb{E}_{\epsilon \sim N(0,1), t}{
        \left[ || \epsilon - \epsilon_\phi(v_\theta(y)_t, t) ||_2^2 \right]
    },
\end{equation}
as illustrated in \cref{fig:dm_training}~(right).
The embedding mapping process of the prior model uses the frozen pretrained vocabularies from the layout-conditional model,
and the architecture of the denoising network here is a transformer~\cite{vaswani2017attention}, which is similar to the layout encoder in the layout-conditional model.

With the learned prior and the pretrained conditional model, theoretically, we can accurately compute the maximum of Bayes' posterior according to Eq.~\eqref{eq:gen_to_dis} to create an object detector.
Another challenge is that there are countless potential detection labels for a given test image.
For one thing, the number of objects is undetermined.
Moreover, even for a single object, the number of possible bounding boxes on the image increases quadratically with respect to its size.
Therefore, a non-enumeration-based approach is needed.

\subsubsection{Optimization-based Diffusion Model Inversion}
Rather than trying the countless possible predictions and then choosing the one with the highest posterior value, 
we propose to directly optimize the prediction $y$ by maximizing the posterior objective.
By substituting Eq.~\eqref{eq:cond_dm_train} and~\eqref{eq:label_dm_train} into Eq.~\eqref{eq:gen_to_dis}, we obtain
\begin{multline}\label{eq:original_objective}
    \hspace{0.1cm} y^* = \argmin_{\redText y}\big\{
        \mathbb{E}_{\epsilon \sim N(0,1), t}{
            \left[ || \epsilon - \epsilon_\theta(x_t, t, v_\theta({\redText y})) ||_2^2 \right]
        }\\
        +
        \mathbb{E}_{\epsilon \sim N(0,1), t}{
            \left[ || \epsilon - \epsilon_\phi(v_\theta({\redText y})_t, t) ||_2^2 \right]
        }
    \big\},
\end{multline}
where we simply add the losses of the two diffusion models.

A practical issue is that the embedding mapping (vocabulary lookup) operation $v_\theta(\cdot)$ is not differentiable.
Therefore, we turn to the embedding space to deploy our learnable parameters $v$,
and restrict them to lie in the discrete output space of $v_\theta(\cdot)$, 
by using the nearest embeddings in corresponding vocabularies $\textit{NN}_\text{value}(v)$ to replace these learnable parameters $v$.
Furthermore, during backward propagation, we copy the gradients from the replacement $\textit{NN}_\text{value}(v)$ to the original learnable parameters $v$. 
This discrete optimization trick enables the gradient to flow back to our learnable parameters, similar to the training of discrete autoencoders~\cite{van2017neural, razavi2019generating, esser2021taming, ramesh2021zero}.

Formally, Eq.~\eqref{eq:original_objective} is implemented as:
\begin{multline}\label{eq:objective}
    \hspace{0.68cm} v^* = \argmin_{\redText v}\big\{
        \mathbb{E}_{\epsilon \sim N(0,1), t}{
            \left[ || \epsilon - \epsilon_\theta(x_t, t, v') ||_2^2 \right]
        }\\
        +
        \mathbb{E}_{\epsilon \sim N(0,1), t}{
            \left[ || \epsilon - \epsilon_\phi({v'}_t, t) ||_2^2 \right]
        }
    \big\},
\end{multline}
\begin{equation}\label{eq:discrete_optim}
\hspace{-2.58cm} v' := \mathrm{sg}\left[\textit{NN}_\text{value}({\redText v}) - {\redText v}\right] + {\redText v},
\end{equation}
where $\mathrm{sg}$ is the stop-gradient operator, and $\textit{NN}$ represents the nearest neighbor, using cosine distance as the default metric.
The whole framework is illustrated in \cref{fig:dm_inv}.

After optimization, the final prediction $y^*$ can be obtained by finding the indices of the nearest embeddings of $v^*$ in the pretrained vocabularies:
\begin{equation}\label{eq:get_result}
    y^* = \textit{NN}_\text{index}(v^*).
\end{equation}

\subsubsection{Implementation}
\label{sec:method_impl}
\paragraph{Calculation of the Expectations}
When learning $v$ using Eq.~\eqref{eq:objective} at each step, the noises and timesteps are always sampled from the corresponding distributions, similar to the training procedure of the diffusion models.
We update $v$ for a number of optimization steps.
To obtain $v^*$, we cannot directly compare the loss values of each update because their noise and timestep values differ. 
Therefore, we pre-saved a set of fixed noises and timesteps and use them to periodically evaluate the loss value during the optimization process. 
These values serve as a monitor for selecting the $v^*$. 
All the results reported in \cref{sec:exp} are obtained in this way.

\paragraph{Other Details}
We empirically find in our preliminary experiments that initializing the learnable embeddings $v$ with the [none] embeddings in the frozen vocabularies works well on both classification and detection tasks. Thus, we leave this design unchanged.
For the object detection task, when getting detected objects from the optimized sequence, 
we drop none-value-contained objects and illegal objects (i.e., those where $m_{max} \leq m_{min}$ or $n_{max} \leq n_{min}$).
\cref{alg:dive} summarizes the pseudo-code of the optimization process for the object detection task.

\SetAlgoSkip{smallskip}
\setlength{\textfloatsep}{0pt}
\begin{algorithm}[h]
\caption{\label{alg:dive} DIVE algorithm for object detection}
\SetKwFunction{FEval}{eval}
\SetKwProg{Fn}{Function}{:}{}
\KwIn{
    pretrained layout-to-image model $\epsilon_\theta(x_t, t, v)$ with embedding vocabularies $v_\theta: y \to v$, 
    pretrained prior layout model $\epsilon_\phi(v_t, t)$, 
    a set of timesteps $T$ used for inversion (same between the two models),
    test image $x$,
    maximum optimization steps $K$,
    monitoring frequency
}
\KwOut{
    prediction $y$ of the given image $x$
}
\vspace{1mm}
\textcolor{gray}{\# Pre-save two sets of fixed noises for evaluation} \\
$E_{\text{for}\theta} = \{\epsilon_{\text{for}\theta} \sim N(0,1)\}^{range(T)}$ \\
$E_{\text{for}\phi} = \{\epsilon_{\text{for}\phi} \sim N(0,1)\}^{range(T)}$  \\
\vspace{1mm}
\textcolor{gray}{\# Optimization loop} \\
$y = [\text{none}, \text{none}, \cdots, \text{none}]$;\, $v = v_\theta(y)$ \\
$\text{monitor\_values} = []$, $\text{corresponding\_embs} = []$ \\
\For{$k$ in range$(K)$}{
\textcolor{gray}{\# In-vocabulary discrete optimization} \\
$v = \text{stop\_gradient}\left( \textit{NN}_\text{value}(v) - v \right) + v $ \\
$t \sim T$,
$\epsilon_{\text{for}\theta} \sim N(0,1)$, 
$\epsilon_{\text{for}\phi} \sim N(0,1)$ \\
$x_t = \text{addn} (x, t, \epsilon_{\text{for}\theta})$,
$v_t = \text{addn} (v, t, \epsilon_{\text{for}\phi})$ \\
$\mathcal{L} = || \epsilon_{\text{for}\theta} - \epsilon_\theta(x_t, t, v) ||_2^2 + 
               || \epsilon_{\text{for}\phi} - \epsilon_\phi(v_t, t) ||_2^2 $ \\
Update $v$ to minimize $\mathcal{L}$ \\
\textcolor{gray}{\# Periodically evaluate the optimized $v$} \\
\If{$k$ \% monitoring frequency $== 0$}{
    $\text{monitor\_values.append}(\FEval(v))$ \\
    $\text{corresponding\_embs.append}(v)$ \\
    }
}
\vspace{1mm}
\Fn{\FEval{$v$}}{
    $v = \textit{NN}_\text{value}(v)$, drop none and illegal boxes \\
    Calculate val\_losses over $\{T\}$ like \texttt{line10-12} but uses saved fixed noises in $E_{\text{for}\theta}, E_{\text{for}\phi}$ \\
    \KwRet averaged $\text{val\_losses}$ \\
}
\vspace{1mm}
\textcolor{gray}{\# Get final result} \\
$v^* = \text{corresponding\_embs}[\text{argmin}(\text{monitor\_values})]$ \\
\KwRet $y = \textit{NN}_\text{index}(v^*) $, drop none and illegal boxes \\
\end{algorithm}

\section{Experiments}
\label{sec:exp}

We conduct experiments in this section to assess the efficacy of the proposed Diffusion model InVErsion (DIVE) method from the following three aspects:
\begin{itemize}[leftmargin=*]
    \item Performing object detection through using diffusion model only (\cref{sec:exp_det}).
    \item Speeding up previous Diffusion Classifier method~\cite{li2023your} on image classification task (\cref{sec:exp_cls}).
    \item Providing a self-contained evaluation metric for conditional diffusion models (\cref{sec:exp_eval}).
\end{itemize}

\subsection{DIVE for Object Detection}
\label{sec:exp_det}

In this subsection, we aim to answer the following questions to evaluate our method for the standard object detection task:
\begin{itemize}[leftmargin=*]
    \item How does our diffusion model inversion method compare to other diffusion-model-based methods and purely discriminative methods?
    \item How do the two proposed components, prior modeling and discrete optimization, affect the final result?
\end{itemize}

\subsubsection{Setup}

\paragraph{Dataset and Evaluation Protocol}
We conduct experiments using the object detection dataset COCO~2017~\cite{lin2014microsoft}, containing 118k training and 5k validation images. 
Although our optimization-based method has already made performing detection implementable, 
due to the still high compute cost of generative methods, 
we evaluate DIVE on a subset of 500 images.
We report average precision (AP), a metric averaged over multiple Intersection over Union (IoU) thresholds, which is regarded as the single most important metric when evaluating performance on COCO~\cite{lin2014microsoft}.
Other metrics (AP at single thresholds of 0.5 and 0.75, and on objects of small, medium, and large sizes) are also reported for thoroughly analyzing our method, following the definitions in~\cite{lin2014microsoft}.

\paragraph{Pretrained Layout-to-Image Diffusion Model}
Since there is no publicly available pretrained layout-to-image diffusion model on the COCO object detection dataset\footnote{Some works~\cite{rombach2022high, fan2023frido, cheng2023layoutdiffuse, zheng2023layoutdiffusion, chen2024geodiffusion} train layout-to-image models on COCO-Stuff dataset~\cite{caesar2018coco} (sometimes dubbed COCO), but that dataset is not suitable for the object detection task~\cite{caesar2018coco}.
Some works~\cite{yang2023reco, li2023gligen} extend pretrained text-to-image models to support positional inputs, but inverting such models is not suitable for the standard detection task because they involve captions.},
we retrain a model using the official code from LDM~\cite{rombach2022high}. 
The denoising network adopts UNet~\cite{ronneberger2015unet} architecture, and the layout encoder is a transformer~\cite{vaswani2017attention}.
Following its implementation, we train the model in the latent space of VQGAN~\cite{esser2021taming} with a downsampling factor of 8 (named LDM-8).
The model is trained at an image resolution of $256\times256$, and the input layout sequence has a fixed padding length of 100 objects.
Object coordinates are kept in integer precision, meaning there are $256+1$ embeddings stored in the bounding box vocabulary.
We add sequence-length learnable positional encodings to the embedded layout (box and class) sequence. During training, the objects in an image are randomly shuffled in different iterations since they should not have a fixed order, i.e., objects are in equivalent positions in the sequence.
Such a model, with other designs unchanged, has a size of 363M parameters.
We train the model over 200k iterations ($\sim$220 epochs), taking about 5 days with 8 Nvidia A100 GPUs.

\begin{table}[t]
\caption{\textbf{Influence of optimization step of DIVE.}
The speed is evaluated on a Nvidia 3090 GPU.
We observe a $\sim$1.8 speedup ratio if switching to a more advanced A100 GPU, as summarized in \cref{sec:ablat_gpu}.
}
\label{tab:optim_steps}
\vskip -0.06in
\centering
\scalebox{1.06}{
    \begin{tabular}{c|cccc}
    \toprule
    Optimization step & 400 & 1000 & 2000 & 4000 \\
    Time per image     & 18min & 45min & 1.5h & 3h \\
    AP                 & 5.7 & 6.4 & 7.1 & \textbf{7.2} \\
    \bottomrule
    \end{tabular}
}
\vskip -0.11in
\end{table}

\begin{figure}[t]
  \centering
  \includegraphics[width=.93\linewidth]{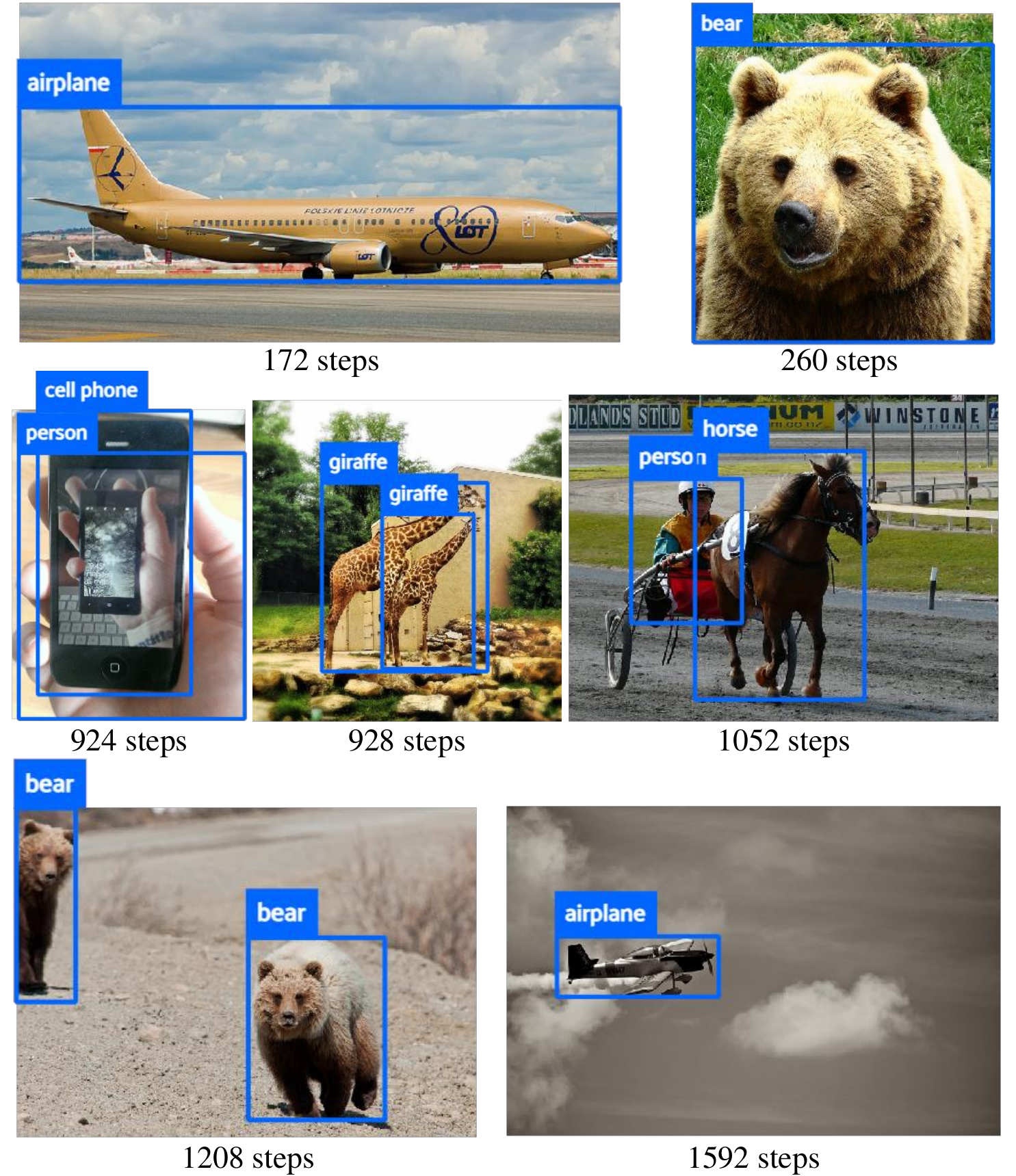}
  \vskip -0.04in
  \caption{\textbf{Visualization of DIVE detection results with corresponding convergence optimization steps shown below,}
  yielded from the monitor introduced in \cref{sec:method_impl}.
  The maximum optimization step for all images is set as a fixed number (2000) here for simplicity.
  The average convergence step of the test set is about 960.
  Images fed to the networks are in $256\times256$ resolution and shown here in their original aspect ratios for better visualization.
  }
  \label{fig:det_vis_w_steps}
  \vskip 0.02in
\end{figure}

\vskip 0.05in
\paragraph{Prior Layout Diffusion Model}
The prior layout diffusion model shares a similar architecture with the layout encoder in the layout-to-image model.
Both are transformers, and we simply inherit the implementation, resulting in a model size of 58M.
The differences lie in their function - in the prior model, the transformer serves as the denoising network, while in the layout-to-image model, it functions as the conditional encoding branch.
We train the prior model over 500k iterations ($\sim$270 epochs), taking about 3 days with 4 Nvidia 3090 GPUs.

\vskip 0.05in
\paragraph{DIVE Details}
We inherit the original LDM's hyper-parameter choices in their code when using the trained LDM to optimize our embeddings, unless stated otherwise.
Following \cite{li2023your}, we use evenly spaced timesteps for improving learning and monitoring efficiency. The even interval is 5, thus, we have 200 different timesteps $\{2, 7, \cdots, 997\}$.
The monitor activates after a loop over these different timesteps.
We conduct experiments with a batch size of 50, consisting of 5 steps of gradient accumulation due to the GPU memory limit.
The learning rate is set to 0.01 with AdamW optimizer~\cite{loshchilov2019decoupled}.
We use a single Nvidia 3090 GPU to invert each image.
The computational cost with respect to the optimization step of DIVE is discussed in the next paragraph.

\paragraph{DIVE Optimization Step and Computational Cost}
\label{sec:ablation_optim_steps}
We test a group of optimization steps to assess its influence.
\cref{tab:optim_steps} shows the quantitative results, where all images use the same optimization step in each trial.
We do not observe significant AP gains when the steps exceed 2000.
We note that the computational cost is relatively higher than common discriminative detectors since DIVE requires several times of backward propagation through the whole network to optimize the result.
Although the computational cost of the detection task is currently not the main focus of this work,
which is to make the detection task doable using pretrained frozen image generation models and check their discriminative ability,
the high cost of DIVE pushes us to explore whether there is potential for reduction without compromising the detection performance.

\vskip 0.03in
To this end, we check the convergence step of some images and show them in \cref{fig:det_vis_w_steps} together with the detection visualization results.
It can be seen that for some ``easier'' test images with larger and fewer objects, the optimization process converges more quickly.
For ``harder'' ones, DIVE requires more steps to converge.
Based on this observation, we can conclude that using an adaptive optimization-step strategy would be a feasible way to reduce the overall computational cost while preserving detection performance.
For example, we can use an early stopping strategy to halt the optimization process if the monitor value (loss) does not decrease for a number of steps.
We further check the average convergence step of the test set, which is about 960, indicating that the early stopping method could reach a maximum of $\sim2\times$ speedup ratio overall.
Currently, for simplicity, the reported DIVE results of the following experiments are obtained using a fixed 2000 steps of optimization for all images.

\vskip 0.07in
\subsubsection{Comparison to Discriminative Detectors}

\paragraph{Baselines}
We first compare our \textit{purely generative} method to \textit{generative-discriminative-mixed} methods that also leverage a pretrained image-generation diffusion model.
\textbf{Synthetic Data} trains a discriminative detection model using synthetic training data generated from the layout-to-image model.
\textbf{Diffusion Feature} uses the pretrained diffusion model to extract features, serving as the backbone in discriminative object detectors.
These baselines are essentially discriminative approaches, and we use Faster R-CNN~\cite{ren2015faster} as their backend.
For the \textbf{Diffusion Feature} method, in addition to LDM-8, we also use the text-to-image model Stable Diffusion~\cite{rombach2022high} (SD v1.5), which is trained on a web-scale dataset much larger than COCO, serving as the feature-extraction module, following~\cite{zhao2023unleashing}.

We further compare with widely recognized \textit{purely discriminative} object detectors \textbf{Faster R-CNN}~\cite{ren2015faster} and \textbf{DETR}~\cite{carion2020end},
and a more recent method, \textbf{DiffusionDet}~\cite{chen2023diffusiondet} (\textbf{DiffDet~$S$~@~$N_{eval}$}), that deploys the diffusion process in the proposal boxes space based on discriminative image features, where $S$ and $N_{eval}$ are two hyper-parameters introduced in the paper~\cite{chen2023diffusiondet}.
We adopt various backbones (ResNet-50 and ResNet-101~\cite{he2016deep}) and optimized architectures (feature pyramid network (FPN)~\cite{lin2017feature} and dilated last stage (DC5)~\cite{li2017fully}) for these detectors.
We use \texttt{detectron2}~\cite{wu2019detectron2} to implement Faster R-CNN and DiffDet baselines and use the official code to implement DETR baselines.
The Faster R-CNN and DiffDet models are trained using the 3$\times$ schedule ($\sim$37 epochs), which is the typical setting in \texttt{detectron2}.
DETR models are trained for over 200 epochs, as transformer-based architectures generally require longer training schedules.
These discriminative detectors are trained at a resolution of $256\times256$, using random cropping and horizontal flip augmentations, and are trained from scratch, consistent with the layout-to-image diffusion model setup.
Although it is difficult to make truly fair comparisons due to differences in architectures and training costs, our comparisons aim to thoroughly evaluate the inverted diffusion model regarding its strengths and weaknesses.

\begin{table}[t]
    \caption{\textbf{Object detection on COCO validation set.} We compare to generative-discriminative hybrid methods that also use image generation diffusion models. We also compare to advanced purely discriminative methods.
    T stands for time per image, which is evaluated on a Nvidia 3090 GPU.
    }
    \label{tab:det}
    \vskip -0.03in
    \centering
    \setlength{\tabcolsep}{.9mm}
    \scalebox{1.05}{
    \hskip -0.08in
    \begin{tabular}{llccccccc}
    \toprule
    Method & Backbone & T & AP & \AP{50} & \AP{75} & \AP{S} & \AP{M} & \AP{L} \\
    \midrule
    \multicolumn{8}{l}{\textit{Generative + discriminative methods:}} \\
    Synthetic Data      & R50   &   90ms   &   4.7	    &   8.7	    &   4.5	    &   0.0	&   3.1	  &  9.8    \\
    Diffusion Feature   & LDM-8 &   47ms   &   6.9	    &   13.4	&   5.9	    &   0.0	&   3.5	  &  15.8   \\
    Diffusion Feature & SD & 41ms  & 7.9 & 15.7 & 7.3 & 0.0 & 1.2 & 18.9 \\
    \midrule
    \multicolumn{8}{l}{\textit{Generative methods:}} \\
    \textbf{DIVE (ours)} & LDM-8 & 1.5h  &   7.1     &   11.0    &   7.1     &   0.0 &   1.2   &  16.4   \\
    \midrule
    \multicolumn{8}{l}{\textit{Discriminative methods:}} \\
    Faster R-CNN & R50         &   90ms &   6.8	    &   12.7	&   5.7	    &   0.3	&   5.0	  &  13.2   \\
    Faster R-CNN & R101-FPN    &   29ms  &   9.6	    &   17.3	&   9.5	    &   2.2	&   9.0	  &  17.2   \\
    DETR        & R50         &   38ms   &   9.3	    &   16.9	&   8.7	    &   1.3	&   6.4	  &  18.3   \\
    DETR        & R101-DC5    &   47ms   &   14.7	    &   24.9	&   13.9	&   2.6	&  11.6	  &  \textbf{27.6}   \\
    DiffDet ($1$@$300$) & R101-FPN & 37ms & 14.6 & 24.4 & 14.6 & 2.9 & 13.8 & 25.9 \\
    DiffDet ($4$@$500$) & R101-FPN & 41ms & \textbf{15.2} & \textbf{25.7} & \textbf{14.8} & \textbf{3.4} & \textbf{14.4} & 26.9 \\
    \bottomrule
    \end{tabular}
    }
    \vskip 0.07in
\end{table}

\vskip 0.03in
\paragraph{Results}
\cref{tab:det} shows the comparison results\footnote{The relatively lower reproduced results of discriminative detectors is because we use a smaller image size of $256\times256$, which is common for generative models, for fair comparisons.}.
Compared to other generative-discriminative-mixed methods that use the same pretrained diffusion model,
DIVE outperforms Synthetic Data and is competitive with Diffusion Feature (LDM-8).
The corresponding visualization results of these methods are displayed in the left part of \cref{fig:det_ablation_vis}.
The Diffusion Feature baseline equipped with the SD backbone performs better than our DIVE and other LDM-8-based methods.
This is because the text-to-image SD was pretrained on a larger dataset than LDM-8's COCO, 
incorporating richer prior knowledge into the network.

Compared to purely discriminative detectors, 
DIVE achieves competitive results against the basic discriminative detector Faster R-CNN R50, 
but falls behind stronger detectors that use deeper backbones or modern architectures.
Notably, to the best of our knowledge, we are the first to show that using a frozen pretrained image generation model can successfully tackle the challenging object detection task and achieve competitive performance compared to discriminative methods.

Furthermore, we conduct a deeper analysis of DIVE by comparing it to Faster R-CNN R50, which achieves a competitive average performance.
An interesting phenomenon is that, although falling behind Faster R-CNN R50 under \AP{50},
DIVE performs better under \AP{75} and slightly wins overall AP.
This phenomenon indicates that DIVE has higher precision in true positive bounding box predictions while leaving some objects undetected.
These undetected objects are usually small and medium objects, as indicated by poorer \AP{S} and \AP{M} values.
Not coincidentally, the Synthetic Data and Diffusion Feature baselines also suffer from this issue.
We conjecture that this is because LDM and SD are trained in the latent space, which has a downsampling factor of 8.
In the compressed modeling space, smaller objects are resized to be even smaller.

\vskip 0.03in
\subsubsection{Ablation Study}
\label{sec:exp:ablation}

\begin{figure*}[t]
  \includegraphics[width=1.02\textwidth]{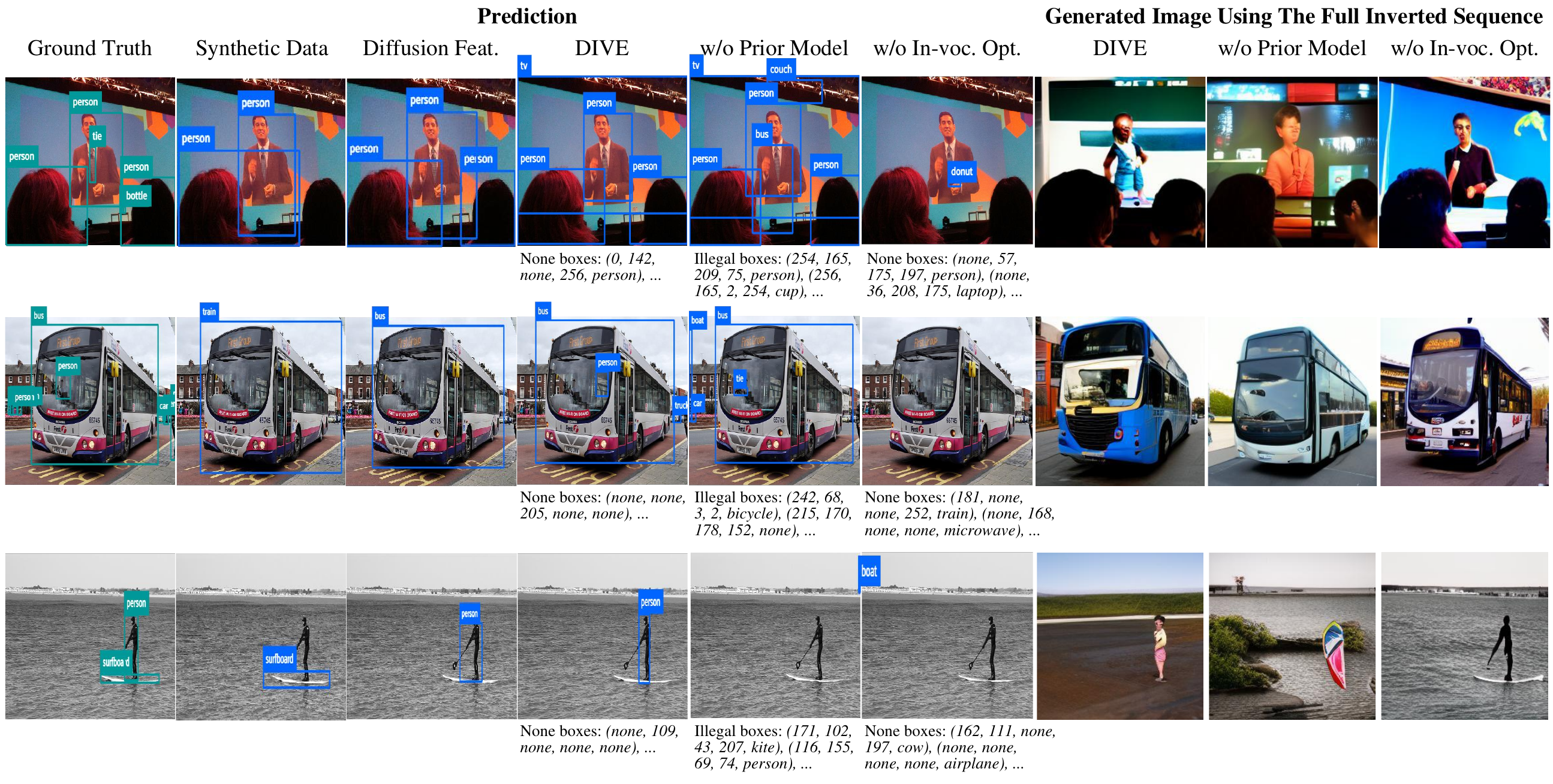}
  \vskip -0.02in
  \caption{\textbf{Visualization of the object detection results.}
  Besides comparing DIVE with other generative baselines that use the same pretrained diffusion model as ours, we also show the influences of the prior model and in-vocabulary discrete optimization method.
  For these ablations, 
  we show at the \textit{bottom} some additional dropped objects (none-value-contained and illegal boxes) in the inverted sequence for clearer visualization of different methods' behavior.
  And on the \textit{right}, we show generated images by feeding the \textit{full} inverted sequence to the pretrained layout-to-image model.
  Zoom in for better visualization.
  }
  \label{fig:det_ablation_vis}
  \vskip -0.16in
\end{figure*}

In this subsection, we study the impact of the proposed components in our method.
Besides ablating the prior model, we also study the impact if the discrete optimization trick were not used, i.e., using Eq.~\eqref{eq:objective} as the objective without the in-vocabulary discrete optimization trick from Eq.~\eqref{eq:discrete_optim} ($v'$ is $v$ in this case).
We conduct the ablation studies on a smaller subset of 100 images.
It can be seen from \cref{tab:det_ablation} that quantitatively, both components are important to the overall performance.
The qualitative results in \cref{fig:det_ablation_vis} tell more about the reason, and we analyze them below.

\vskip 0.02in
\paragraph{Prior Model}
From \cref{fig:det_ablation_vis}, it can be seen that if we do not integrate the prior model into the inversion process, the inverted sequences (predictions) contain many redundant objects and illegal boxes.
The synthetic images tell the reason.
These redundant objects do not hurt the generation process of the conditional model  (e.g., the first two lines),
which is an interesting phenomenon similar to the findings in the literature of attacking text-to-image diffusion models~\cite{milliere2022adversarial, Zhuang_2023_CVPR, gao2023evaluating}. 
However, these redundancies will negatively affect the detection performance, resulting in many false positives. 
Only if the prior model tells the optimization process about which types of layouts are more common could we achieve better detection results (DIVE default).

\begin{table}[t]
    \caption{\textbf{Ablation studies of DIVE on the object detection task.}}
    \label{tab:det_ablation}
    \centering
    \vskip -0.04in
    \setlength{\tabcolsep}{1.8mm}
    \scalebox{1.06}{
    \begin{tabular}{lccc}
    \toprule
    Method & AP & \AP{50} & \AP{75} \\
    \midrule
    DIVE                    & \textbf{10.3} & \textbf{14.0} & \textbf{11.5} \\
    w/o Prior Model         &  6.7 & 10.6 &  6.2 \\
    w/o In-vocabulary Opt.  &  4.6 &  5.6 &  4.8 \\
    w/o both                &  3.2 &  5.4 &  3.2 \\
    Decoupling classes and boxes & 9.0 & 12.4 & 9.2 \\
    \bottomrule
    \end{tabular}
    }
    \vskip 0.08in
\end{table}

\vskip 0.02in
\paragraph{In-vocabulary Discrete Optimization}
In the case of continuous optimization in the embedding space (i.e., w/o using Eq.~\eqref{eq:discrete_optim}), the optimization process only needs to reconstruct (learn the content of) the given image. 
This case can be analogized to Textual Inversion~\cite{gal2023an}, which learns to inject new words for the given images into a pretrained text-to-image model without considering new-word semantics.
For our situation, if we attempt to decode to see what these ``new words'' are (by finding nearest neighbors in the vocabulary), as shown in \cref{fig:det_ablation_vis} right most,
sometimes, we could get correct classes (e.g., the first line ``person''), but in many other cases, classes are difficult to interpret (e.g., the last line ``cow''), and boxes are hard to complete, even with the help from the prior model.
This is because the reconstruction purpose dominates the optimization (best reconstruction performance in \cref{fig:det_ablation_vis}), leaving words' interpretability untouched.

\vskip 0.02in
\paragraph{Decoupling classes and bounding boxes}
Considering that the trained prior model may overfit the training data distribution, this ablation decouples the classes and bounding boxes in the sequence.
Specifically, we train two prior distribution models separately, one for classes and the other for boxes, and use them together during inversion.
However, note that this strategy may not be a natural and proper way to model the prior layout distribution, as it separates the relationships between classes and boxes. 
For example, it cannot distinguish ``a bicycle on a person'' and ``a person on a bicycle'' with the same box layouts.
Nevertheless, it can still learn what boxes are legal compared to not using any prior models.
The last line of \cref{tab:det_ablation} presents the result. 
It falls between the results of the default (un-decoupling) method and the ``w/o Prior Model'' method, which aligns with the above hypothesis.

\subsection{DIVE for Image Classification}
\label{sec:exp_cls}

In this subsection, we compare our optimization-based method with the previous enumeration-based method Diffusion Classifier~\cite{li2023your} on the image classification task regarding accuracy and speed.

\vskip 0.02in
\subsubsection{Setup}
Following the setup in \cite{li2023your}, we use pretrained Diffusion Transformer (DiT)~\cite{peebles2023scalable} as the conditional diffusion model to be inverted, which is a class-conditional model trained on ImageNet-1k~\cite{russakovsky2015imagenet}.
Same as the setup in \cite{li2023your} for a fair comparison, we use DiT-XL/2 at the resolution $256\times256$ and evaluate on a subset of 2000 images (2 images per class).
Given that ImageNet-1k's label distribution is uniform, the prior modeling model is unnecessary in this case.
The number of different timesteps for DIVE here is set to 250 to align with the Diffusion Classier.
We optimize our embeddings over 200 steps with a batch size of 25, taking about 80 seconds to invert an image on a single A100 GPU.

\begin{table}[t]
    \vskip .07in
    \caption{\textbf{Image classification on ImageNet-1k.} 
    The speeds are tested on an A100 GPU.
    The accuracy of Diffusion Classifier is from \cite{li2023your}, and its speed is tested using the official code.
    }
    \label{tab:cls}
    \centering
    \vskip -0.04in
    \setlength{\tabcolsep}{2mm}
    \scalebox{1.09}{
    \begin{tabular}{lcc}
    \toprule
    Method & Time per image &  Accuracy  \\
    \midrule
    \multicolumn{3}{l}{\textit{Generative + discriminative methods:}}\\
    Synthetic Data      &  9 ms    &   67.1   \\
    Diffusion Feature   &  31 ms    &   69.2   \\
    \midrule
    \multicolumn{3}{l}{\textit{Generative methods:}}\\
    Diffusion Classifier    &   $\sim$1100 s    &   77.3    \\
    \textbf{DIVE (ours)}    &    80 s           &   77.2    \\
    \midrule
    \multicolumn{3}{l}{\textit{Discriminative methods:}} \\
    AlexNet   &   8 ms &   57.5 \\
    ResNet-18   &   8 ms    &   70.6    \\
    ResNet-50   &   9 ms    &   77.6    \\
    ResNeXt101  &  15 ms  &  79.7  \\
    ViT-L/32    &   12 ms   &   78.0    \\
    ViT-B/16    &   9 ms    &   81.5   \\
    Swin-B    &    21 ms   &   \textbf{83.7}   \\
    \bottomrule
    \end{tabular}
    }
    \vskip 0.08in
\end{table}

\vskip 0.02in
\subsubsection{Results}
\cref{tab:cls} shows the comparison results.
We compare to Diffusion Classifier~\cite{li2023your}, discriminative classifiers~\cite{he2016deep, dosovitskiy2021an}~\cite{krizhevsky2012imagenet, xie2016aggregated, liu2021swin}, 
and generative-discriminative hybrid methods Synthetic Data and Diffusion Feature.
Synthetic Data trains a standard ResNet-50 using a training dataset generated by DiT-XL/2.
Diffusion Feature, following the baseline design in~\cite{li2023your}, uses the pretrained DiT-XL/2 network as the feature extractor and trains a modified ResNet-18 on top of the features using the ImageNet real dataset.
It can be seen from \cref{tab:cls} that DIVE achieves nearly the same accuracy as the Diffusion Classifier.
This result verifies the effectiveness of the proposed optimization-based method for purely generation-based image classification, representing an ideal alternative to the enumeration-based method.

Notably, our optimization-based method achieves a $\sim14\times$ speedup compared to the enumeration method while maintaining accuracy.
This is because the enumeration method needs to try all possible predictions of the test dataset, which causes a significant computational cost.
Compared to discriminative models that directly map the test image to a prediction label, generative methods still fall behind in efficiency due to the cost of exploring possible labels.
Potential directions for further accelerating these generative methods include reducing the exploration steps or parameters of the pretrained model.

To determine the maximum speedup ratio achievable by using an early stopping strategy to halt DIVE's optimization process, we check the convergence steps of test images here, as done in the previous detection section~\ref{sec:ablation_optim_steps}.
We find that the average convergence step of the test set here is about 106 (max is 200), indicating that reducing the exploration steps could further achieve a maximum of $\sim2\times$ speedup overall.

\vskip 0.02in
\subsubsection{Ablation of In-vocabulary Discrete Optimization}
We ablate the in-vocabulary optimization method here again to verify its effectiveness.
Without this optimization component, we achieve a lower accuracy of 60.3, indicating its importance for the optimization-based approach.

\vskip 0.02in
\subsubsection{Influence of GPU}
\label{sec:ablat_gpu}
Readers may note that we used different GPUs for the previous detection and classification experiments.
There are no other reasons for this but only to schedule our limited resources more efficiently.
Specifically, we tested the speeds of using different GPUs for different tasks before conducting our formal experiments.
As summarized in \cref{tab:ablat_gpu}, we observed that switching from 3090 to A100 achieves a larger speedup ratio on the classification task than on the detection task, leading to the previous choices.
Here, we conduct repeated experiments to assess task performances influenced by switching GPUs.
\cref{tab:ablat_gpu} shows the results, where the detection experiments are conducted at the ablation study's experimental scale (c.f. \cref{tab:det_ablation}). It can be seen that the detection AP and classification accuracy fluctuate slightly.

\subsection{DIVE for Conditional Diffusion Model Evaluation}
\label{sec:exp_eval}

Following \cite{krojer2023diffusion, rambhatla2023selfeval}, 
we use our proposed DIVE to evaluate different conditional diffusion models from the perspective of discriminative ability.
While using DIVE as a general evaluation metric for generative models may be controversial - not applicable to non-likelihood-based models like GANs - it has the advantage of not relying on external models such as the commonly used ImageNet-pretrained InceptionV3~\cite{szegedy2016rethinking} for computing FID~\cite{heusel2017gans}. Nevertheless,
the main aim of this subsection is to see whether the discriminative ability of DIVE aligns with other widely used metrics for evaluating generative models.

\vskip 0.02in
\subsubsection{Setup}
We evaluate both types of conditional diffusion models mentioned earlier.
For class-conditional models, we compare DiT-XL/2~\cite{peebles2023scalable} with LDM-4~\cite{rombach2022high}, where the former, used in the previous subsection, is based on the transformer architecture~\cite{vaswani2017attention}, while the latter is based on the convolutional UNet architecture~\cite{ronneberger2015unet}.
For layout-conditional models, 
we compare the one used in \cref{sec:exp_det} (LDM-8) with another smaller LDM (LDM-8-S) trained ourselves. 
For other evaluation metrics compared, we use Fréchet Inception Distance (FID)~\cite{heusel2017gans}, Inception Score (IS)~\cite{salimans2016improved}, and Precision/Recall~\cite{kynkaanniemi2019improved}.

\vskip 0.02in
\subsubsection{Results}
The results are presented in \cref{tab:eval_metric}.
It can be observed that the results from our discriminative metric align with most other metrics, such as FID, which assesses distribution similarity between real and generated images.
An exception occurs at DIVE vs. Precision in \cref{tab:eval_metric}(a). 
This suggests that a model exhibiting greater generative diversity (high Recall) rather than better image realism (high Precision) may have a stronger discriminative ability (high DIVE).

\begin{table}[t]
    \vskip .07in
    \caption{
    \textbf{Influence of GPU on different tasks.}
    }
    \label{tab:ablat_gpu}
    \centering
    \vskip -0.02in
    \setlength{\tabcolsep}{2mm}
    \scalebox{1.07}{
    \begin{tabular}{cccc}
        \toprule
        Task & GPU & Time per image & AP / Acc \\
        \midrule
        \multirow{2}{*}{\makecell{Detection}} & 3090 & 90 min & 10.3 \\
        & A100 & \textbf{50 min} & \textbf{10.6} \\
        \midrule
        \multirow{2}{*}{\makecell{Classification}} & 3090 & 240 s & 76.6 \\
        & A100 & \textbf{80 s} & \textbf{77.2} \\
        \bottomrule
    \end{tabular}
    }
\end{table}

\begin{table}[t]
    \caption{
    \textbf{Comparing conditional diffusion models with different evaluation metrics.}
    Accuracy and AP are used for DIVE metrics, respectively.
    Other numbers in table (a) are from \cite{peebles2023scalable}.
    Prec. stands for Precision.
    }
    \label{tab:eval_metric}
    \vskip -.04in
    \centering
    \begin{minipage}[c]{\linewidth}
        \centering 
        {\small(a) Comparing class-conditional models}
        \vskip .05in
        \setlength{\tabcolsep}{1.2mm}
        \scalebox{1.07}{
        \begin{tabular}{lcccccc}
        \toprule
        Model & \#Params & FID$\downarrow$ & IS$\uparrow$ & Prec.$\uparrow$ & Recall$\uparrow$ & DIVE$\uparrow$ \\
        \midrule
        LDM-4       &   400M    & 3.60          & 247.67            & \textbf{0.87} & 0.48          & 62.7          \\
        DiT-XL/2    &   675M    & \textbf{2.27} & \textbf{278.24}   & 0.83          & \textbf{0.57} & \textbf{77.2} \\
        \bottomrule
        \end{tabular}
        }
    \end{minipage}
    \vskip .14in
    \begin{minipage}[c]{\linewidth}
        \centering
        {\small(b) Comparing layout-conditional models}
        \vskip .05in
        \setlength{\tabcolsep}{1.4mm}
        \scalebox{1.07}{
        \begin{tabular}{lccccc}
        \toprule
        Model & \#Params & FID$\downarrow$ & Prec.$\uparrow$ & Recall$\uparrow$ & DIVE$\uparrow$ \\
        \midrule
        LDM-8-S & \,\,\,65M & 24.90          & 0.45             & 0.62          & 4.5           \\
        LDM-8   &      363M & \textbf{18.36} & \textbf{0.53}    & \textbf{0.63} & \textbf{7.1}  \\
        \bottomrule
        \end{tabular}
        }
    \end{minipage}
    \vskip .09in
\end{table}

\section{Conclusion}

In this paper, we proposed an optimization-based approach to invert conditional diffusion models to produce discriminative labels under the framework of Bayes' rule.
We empirically show that the proposed optimization-based approach is much faster than the previous enumeration-based method in the image classification task,
and makes it feasible to use purely pretrained generative models for accomplishing the more complex object detection task.
We hope that the conclusion of this paper could help recognize the internal discriminative ability of pretrained image generation models.
Future research could further accelerate the proposed optimization-based approach or extend it to more complex dense tasks, such as semantic segmentation.
However, note that simply extending the proposed approach to dense tasks would create a large optimization space, and the contribution of a single conditional pixel to the final objective is relatively small, which may be problematic and needs to be addressed.
Adding some regularization terms might offer a feasible solution.

\section*{Acknowledgments}
This work is partially supported by National Natural Science Foundation of China (NSFC): 62376259,
62306301, and National Postdoctoral Program for Innovative Talents under Grant BX20220310.

\bibliography{references}
\bibliographystyle{unsrt}

\end{document}